\documentclass[conference]{IEEEtran}
\IEEEoverridecommandlockouts
\usepackage{cite}
\usepackage{amsmath,amssymb,amsfonts}
\usepackage{algorithmic}
\usepackage{graphicx}
\usepackage[subrefformat=parens,labelformat=parens]{subcaption}
\usepackage{textcomp}
\usepackage{xcolor}
\usepackage{url}
\usepackage{latexsym}
\usepackage{makeidx}
\usepackage{moreverb}
\usepackage{tabularx}
\usepackage{verbatim}
\usepackage[capitalize,noabbrev]{cleveref}
\usepackage{rotating}
\usepackage{makecell}
\usepackage{tikz}
\usepackage{array}

\usepackage{tikz}
\usepackage{lipsum} 
\makeatletter
\def\ieeecopyright{ 
    \footnotesize © 2025 IEEE. Personal use of this material is permitted.
    \newline DOI: 10.1109/RTAS65571.2025.00017
} 
\makeatother 
\AddToHook{shipout/firstpage}{% 
    \begin{tikzpicture}[remember picture,overlay] 
        \node[anchor=south west,xshift=1.0cm,yshift=0.8cm] at (current page.south west)% 
        {\parbox{\linewidth}{\raggedright\ieeecopyright}}; 
    \end{tikzpicture}% 
} 
\renewcommand{\arraystretch}{1.5}

\def\BibTeX{{\rm B\kern-.05em{\sc i\kern-.025em b}\kern-.08em
    T\kern-.1667em\lower.7ex\hbox{E}\kern-.125emX}}

\begin{document}

\title{D-AWSIM: Distributed Autonomous Driving Simulator for Dynamic Map Generation Framework}

% \author{Shunsuke Ito, Ryo Okamura, Chaoran Zhao, and Takuya Azumi}
% \affiliation{%
%   \institution{Graduate School of Science and Engineering, Saitama University}
%   \city{Saitama}
%   \country{Japan}
% }

\author{\IEEEauthorblockN{Shunsuke Ito, Chaoran Zhao, Ryo Okamura, and Takuya Azumi}
\IEEEauthorblockA{\textit{Graduate School of Science and Engineering, Saitama University}}
}

\maketitle

\begin{abstract}
    Autonomous driving systems have achieved significant advances, and full autonomy within defined operational design domains near practical deployment. 
    Expanding these domains requires addressing safety assurance under diverse conditions. 
    Information sharing through vehicle-to-vehicle and vehicle-to-infrastructure communication, enabled by a Dynamic Map platform built from vehicle and roadside sensor data, offers a promising solution. 
    Real-world experiments with numerous infrastructure sensors incur high costs and regulatory challenges. 
    Conventional single-host simulators lack the capacity for large-scale urban traffic scenarios. 
    This paper proposes D-AWSIM, a distributed simulator that partitions its workload across multiple machines to support the simulation of extensive sensor deployment and dense traffic environments. 
    A Dynamic Map generation framework on D-AWSIM enables researchers to explore information-sharing strategies without relying on physical testbeds. 
    The evaluation shows that D-AWSIM increases throughput for vehicle count and LiDAR sensor processing substantially compared to a single-machine setup. 
    Integration with Autoware demonstrates applicability for autonomous driving research.
\end{abstract}


\section{Introduction}\label{sec:Introduction}\label{sec:Introduction}
    Current autonomous driving systems are capable of operating without human input and are fully autonomous within operational design domains (ODDs).
    Expanding the ODDs necessitates addressing various challenges, with safety enhancement remaining a paramount priority.
    An effective approach to achieving the desired safety level is the sharing of information available from vehicles or between vehicles and infrastructure sensors, known as Vehicle-to-Vehicle (V2V) or Vehicle-to-Infrastructure (V2I) communications~\cite{cyo-IV}.
    Infrastructure sensors, including LiDARs~\cite{RAGE_nogu} and cameras~\cite{RecognitionByCamera}, are installed along roadways to detect objects and share the recognized data with nearby vehicles.
    By utilizing this shared data, vehicles can anticipate hazards such as blind spots and execute appropriate control strategies.
    However, direct peer-to-peer wireless communication can introduce complexities in communication management~\cite{ComplexityOfV2X} and diminish data reusability.
    
    One promising approach for managing data from infrastructure sensors and vehicles is the deployment of an information platform termed the Dynamic Map (DM)~\cite{DynamicMapAbs}, which builds upon the concept of the Local Dynamic Map (LDM)~\cite{LDMSurvey} originally proposed by ETSI~\cite{LDMDefinesByETSI}.    
    Infrastructure sensors deployed at roadside locations often operate with limited computational resources, necessitating energy-efficient operation~\cite{InfraSensorEnergy}.
    To address these limitations, the Dynamic Map adopts a three-layer architecture.
    This three-layer architecture alleviates the burden on local processing, reducing communication latency.
    
    While the Dynamic Map offers a robust solution for integrating and leveraging data, its adoption for research and development faces practical barriers.
    In particular, the installation of infrastructure sensors necessitates authorization from public authorities, thereby restricting the organizations permitted to install such sensors on public roads. 
    Even with these permissions, modifications to real-world environments are often infeasible, consequently limiting the flexibility essential for research. 
    These constraints impede the broader adoption of the Dynamic Map in research contexts.
    Challenges that are difficult to address in real-world environments are commonly tackled using a simulator.
    However, simulating large-scale urban environments on a single machine is challenging due to processing load and other constraints.
    Therefore, a simulator capable of distributed processing across multiple machines with scalability is required.
    
    This paper proposes D-AWSIM, a simulator designed to enable anyone to freely utilize the Dynamic Map by distributing processing across multiple interconnected machines, thereby enhancing scalability.
    By using D-AWSIM, the computational load of the simulator can be distributed among machines, allowing various types of data, such as sensor data and vehicle data, to be extracted while maintaining simulation accuracy.
    The extracted simulation data is aggregated on a server for stream processing, which is flexibly processed in real time using an SQL-based language.
    
    The contributions of this paper are as follows:
    \begin{itemize}
        \item D-AWSIM enables distributed simulation using multiple machines, ensuring scalability while facilitating the generation of a Dynamic Map without a real-world environment.
        \item The proposed framework enables the free deployment of numerous infrastructure sensors on distributed simulators, facilitating the acquisition of infrastructure sensor data in flexible and customizable traffic environments.
        \item The proposed framework enhances the applicability of the Dynamic Map to a wide range of research and development use cases.
    \end{itemize}

    The remainder of this paper is organized as follows.
    Section~\ref{sec:System-Model} presents an overview of D-AWSIM and a Dynamic Map.
    Section~\ref{sec:Design} describes the details of D-AWSIM, and Section~\ref{sec:Evaluation} evaluates D-AWSIM.
    Section~\ref{sec:lesson_learned} discusses the lessons learned.
    Section~\ref{sec:Related} describes related work, and a brief conclusion is given in Section~\ref{sec:Conclusion}.

\section{System Model}\label{sec:System-Model}
    An overview of D-AWSIM, the Dynamic Map, and the Dynamic Map generation framework is provided to facilitate comprehension of the proposed method details.

    \subsection{Overview of D-AWSIM}\label{subsec:Overview-of-D-AWSIM}
        \begin{figure}[t]
            \centering
            \includegraphics[width=\linewidth]{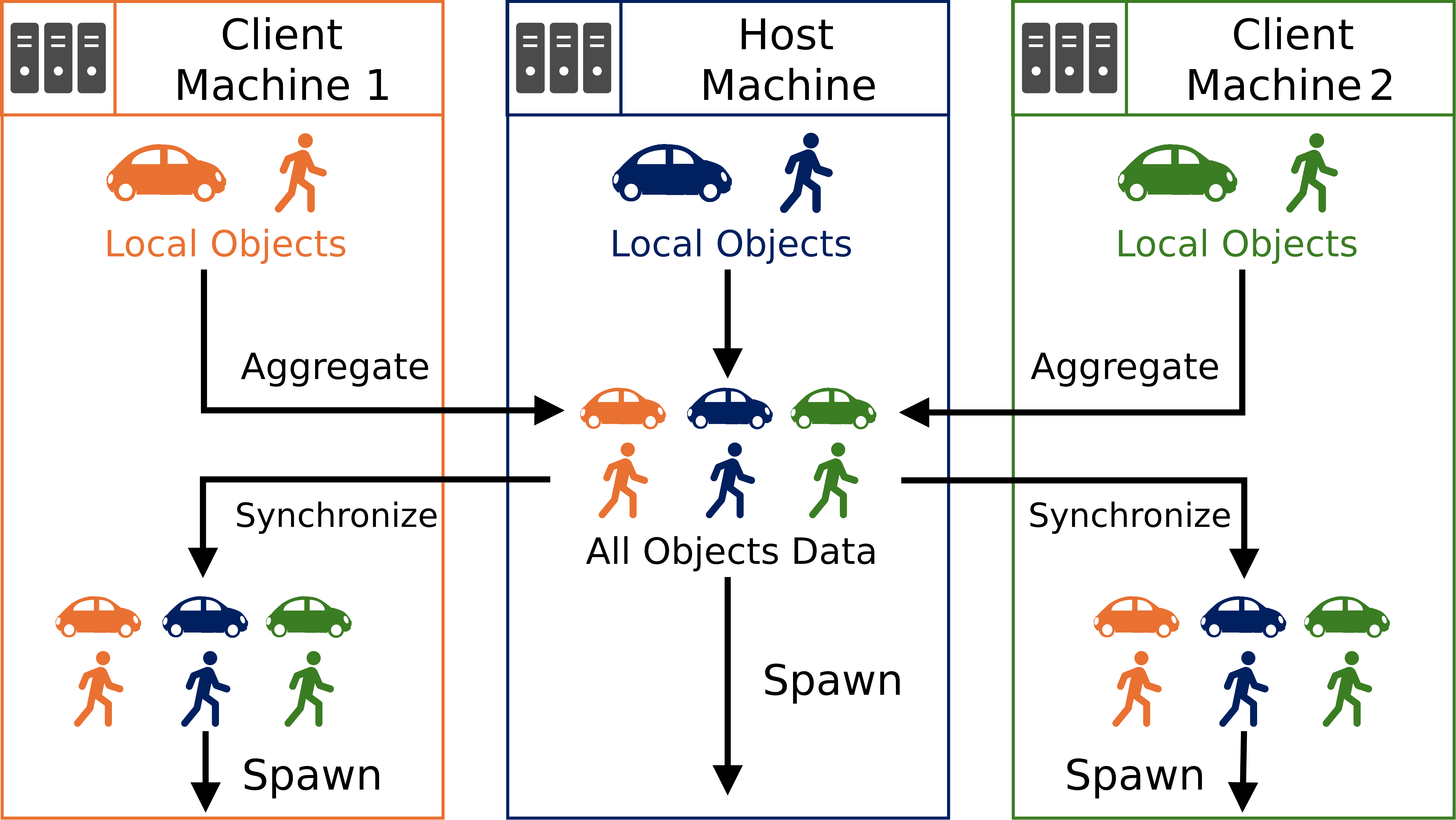}
            \caption{Overview diagram illustrating D-AWSIM interconnection flow.}
            \label{fig:SysModel_D-AWSIM}
        \end{figure}
        \begin{figure}[t]
            \centering
            \includegraphics[width=\linewidth]{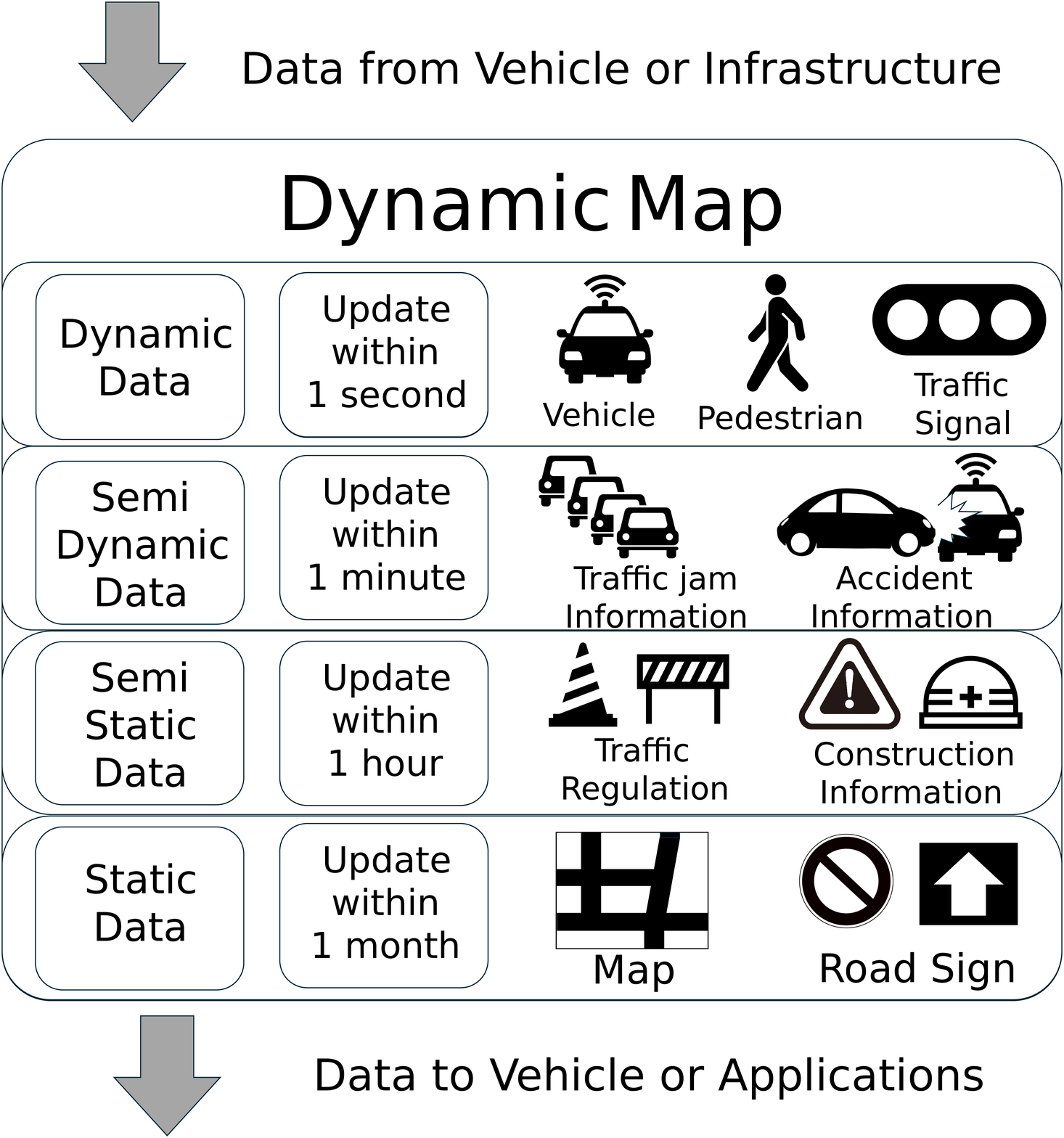}
            \caption{Handled data in the Dynamic Map.}
            \label{fig:DynamicMap_Layer}
        \end{figure}
    
        An overview diagram of D-AWSIM is shown in Fig.~\ref{fig:SysModel_D-AWSIM}. 
        D-AWSIM is divided into one host machine and the remaining client machines. 
        The host machine functions as both server and client responsible for the management of the interconnection. 
        The object synchronization across machines is achieved by aggregating the object data managed by each client on a host, which then redistributes the aggregated data back to all clients.
        Each machine generates objects in its local environment based on the received aggregated data.
        The details of this process are described in Section~\ref{sec:Design}.

        D-AWSIM is a combination of AWSIM, which is an open-source autonomous driving simulator, and Netcode for GameObjects (NGO), a networking library intended for multiplayer games.
        AWSIM is a simulator developed using Unity and features a broad range of sensors, such as LiDARs and cameras, as well as objects including vehicles and pedestrians. 
        The deployment of NGO, provided by Unity Technologies, facilitates the interconnection of AWSIM across multiple machines to construct one large-scale simulation environment.

    \subsection{Dynamic Map}\label{subsec:DynamicMap}
        The Dynamic Map is based on the Local Dynamic Map (LDM) defined by ETSI, and originates from a platform defined by SIP-adus as indispensable infrastructure for autonomous driving vehicles~\cite{SIPAdus-DynamicMap}. 
        The Dynamic Map is layered according to update frequency and type of information, as illustrated in Fig.~\ref{fig:DynamicMap_Layer}, where embedded nodes such as vehicles and infrastructure sensors correspond to the dynamic data.
        
        The Dynamic Map adopts high-speed data processing using stream processing and in-memory management to process data that flow continuously and nearly permanently from embedded nodes. 
        Stream processing differs from batch processing by processing data immediately upon arrival. 
        In-memory management retains and processes data in main memory (RAM) rather than on disk. 
        A data stream management system (DSMS) performs these functions.
        Moreover, the geographic distribution of edge nodes and computation offload functionality are adopted to address high computational load and minimize communication latency.
        
        While edge nodes handle information within a defined area, cloud nodes aggregate information from edge nodes and process data over a broader region. 
        The main roles of cloud servers include enabling large-scale operations. 
        These operations, such as optimizing traffic flow across the entire city and predicting traffic congestion, are facilitated through communication with edge servers. 
    
    \subsection{Generation of Dynamic Map using D-AWSIM}
        \begin{figure}[t]
            \centering
            \includegraphics[width=\linewidth]{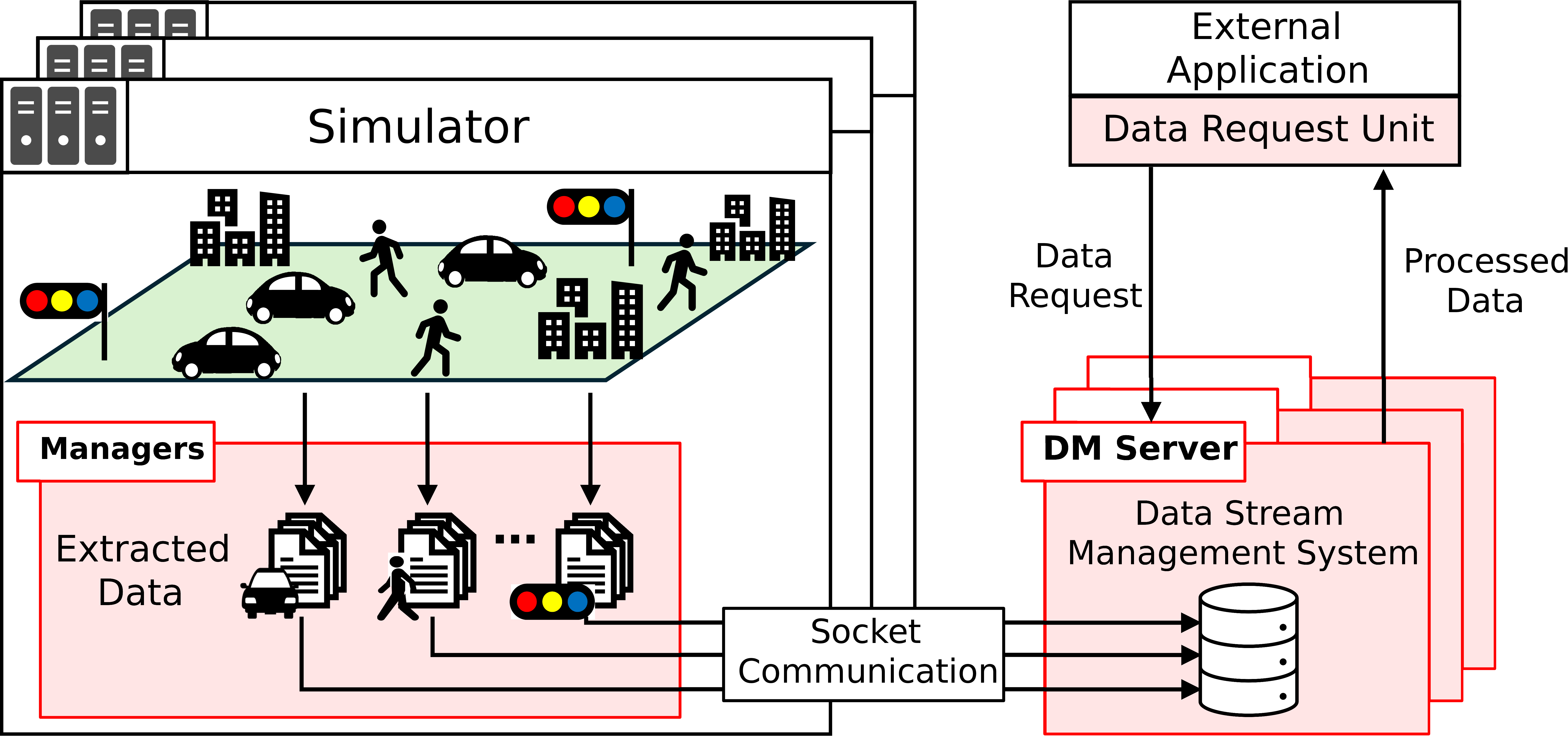}
            % \vspace{2mm}
            \caption{Overview of generation of Dynamic Map.}
            % \vspace{2mm}
            \label{fig:SysModel_DM}
        \end{figure}
        The generation of a Dynamic Map using D-AWSIM proceeds as depicted in Fig.~\ref{fig:SysModel_DM}. 
        The integration of the DM managers for extraction of individual data from objects such as vehicles and pedestrians existing in D-AWSIM is performed. 
        The managers asynchronously extract data from objects on D-AWSIM and aggregate them into a single dataset. The aggregated data is transmitted via socket communication to the DM Server, which functions as DSMS. 
        The DM Server performs stream processing based on SQL-based queries received externally and returns processing results. 
        Communication via HTTP/HTTPS enables interconnection with various systems.

\section{Design and Implementation}\label{sec:Design}
    The detailed description regarding the interconnection of D-AWSIM with other machines and the generation of a Dynamic Map is provided.

    \subsection{Mechanism of D-AWSIM Interconnection}\label{subsec:Mechanism}
        \begin{figure}[t]
            \centering
            \includegraphics[width=\linewidth]{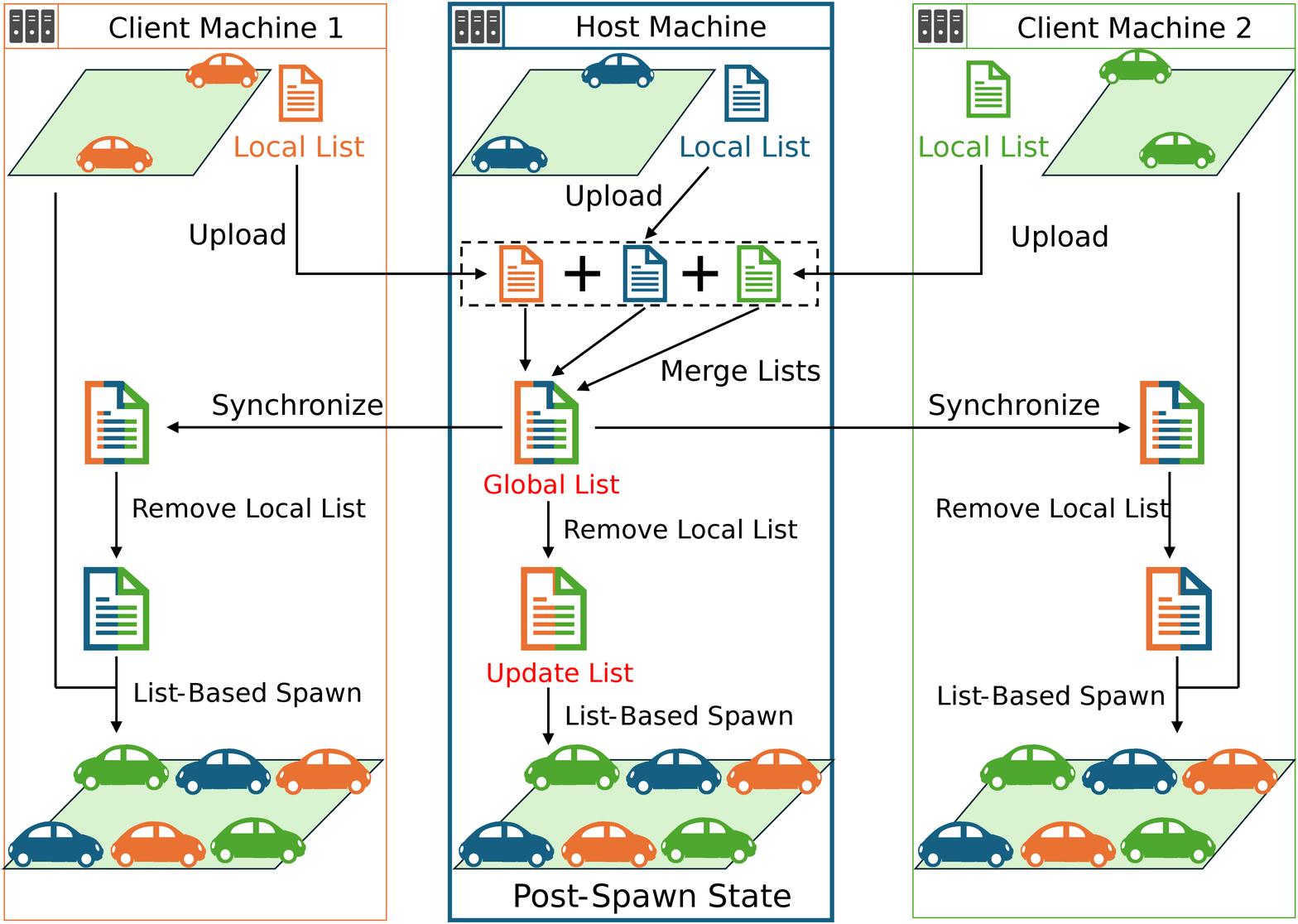}
            % \vspace{2mm}
            \caption{Overview diagram illustrating D-AWSIM interconnection flow.}
            % \vspace{2mm}
            \label{fig:Detailed_D-AWSIM}
                \vspace{-5mm}
        \end{figure}
        
\begin{figure}[t]
            \centering
            \includegraphics[width=\linewidth]{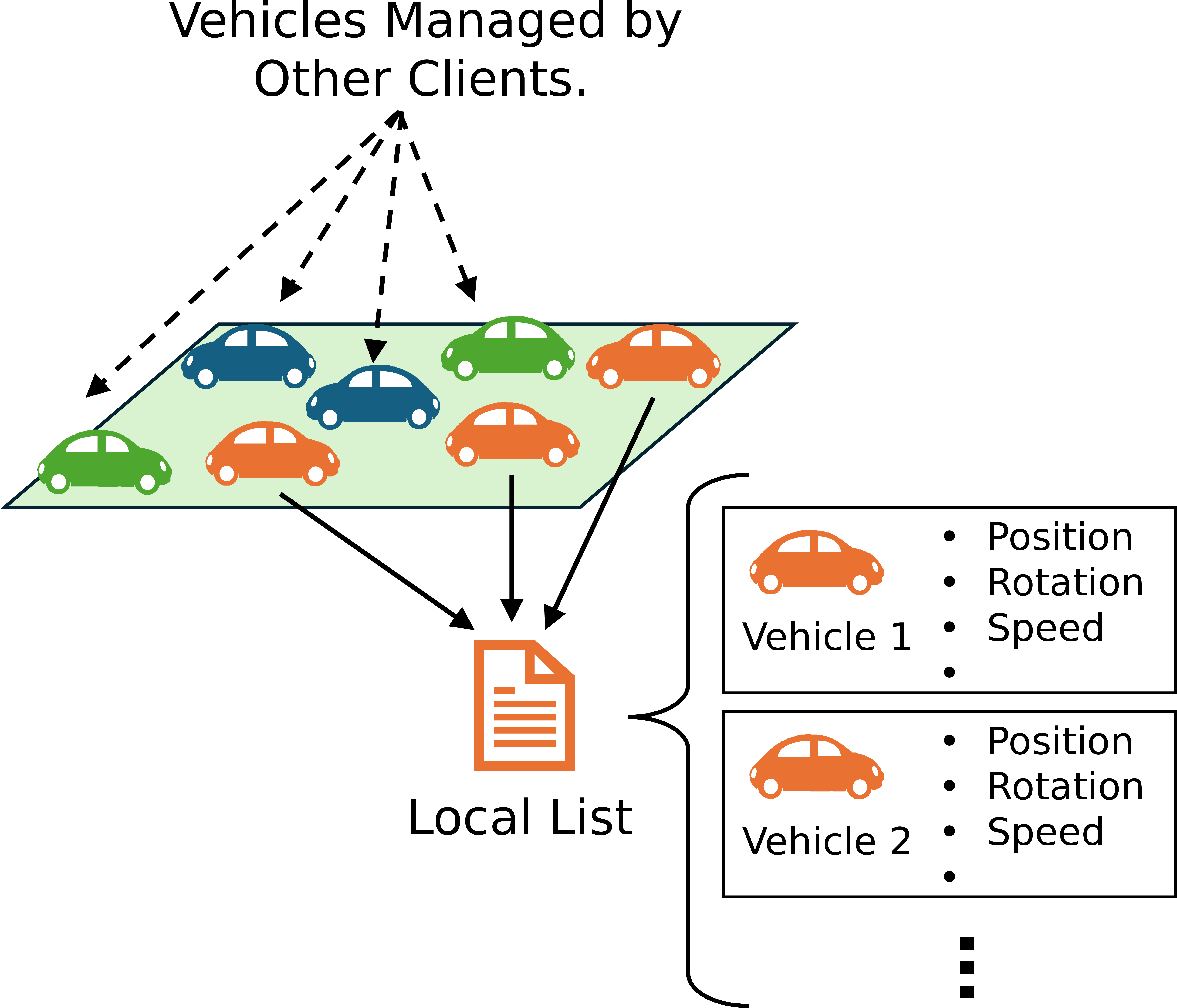}
            % \vspace{2mm}
            \caption{Overview of the local list.}
            % \vspace{2mm}
            \label{fig:Local_List}
            \vspace{-4mm}
        \end{figure}

        D-AWSIM synchronizes objects across machines using local lists, a global list, and update lists, as illustrated in Fig.~\ref{fig:Detailed_D-AWSIM}.
        The local list, as illustrated in Fig.~\ref{fig:Local_List}, is a list stored on the local environment that contains vehicle data managed by the client on each client machine. 
        The stored data, necessary for synchronization, include vehicle position, rotation, and the client ID for management. 
        The client ID is an ID uniquely assigned to each client and is used to identify which client manages a synchronization target object.

        After updating the local list, updating the global list is performed.
        The global list is an aggregation of the local lists held by each client and contains information on the synchronized objects at the time of aggregation.
        Each client employs Remote Procedure Call (RPC) to add the contents of its local list to the global list maintained by the host.
        RPC is a mechanism to execute code on a remote machine via networks, which enables a client to request that a server execute processes, such as synchronization among clients, that can only be performed by a server.
        Each client serializes the information required for RPC, including the local list, and transmits an RPC packet to the host.
        Upon successful authentication on the host, the specified process is executed on the host.
        In this manner, the dynamic update of the global list by each client is enabled.
        After the global list is updated, the updated global list is automatically broadcast to all clients.

        Each client compares the update list from the previous cycle used for vehicle synchronization with the received global list and updates the update list accordingly.
        First, each client references the client ID assigned to each vehicle in the global list and skips vehicles managed by the client itself.
        Next, if a vehicle exists in the previous update list, the vehicle information in the update list is updated with new information.
        Then, if a vehicle in the global list is not present in the update list, that vehicle is registered and added to the update list as a synchronization target object.
        Finally, if a registered vehicle exhibits no information update over a predetermined number of frames, the vehicle information is deleted, and the vehicle is removed from synchronization targets.
        Using the final list obtained through this process, vehicle generation, update, and deletion are performed.

    \subsection{Method for Integrating Additional Machine}\label{subsec:IntegratingMachine}
        All connections between a host machine and client machines are established using Unity Transport, a UDP-based communication method.
        The host machine is launched first, and each subsequent machine connects to the host as a client.
        The host is configured to accept all incoming client connections without any special authentication.
        Therefore, additional machines can be connected without any special configuration, even when scaling up the system.
        
        D-AWSIM can theoretically coordinate an unlimited number of machines and distribute processing tasks accordingly.
        The first client machine to connect is assigned not only the tasks performed by other client machines, but also the processing of the camera mounted on the Ego vehicle managed by the host.
        This is because the Ego vehicle is equipped with sensor components such as LiDAR, cameras, and GNSS.
        Among these, LiDAR and camera processing are particularly demanding and can become performance bottlenecks.
        To address this, LiDAR and camera processing are offloaded to separate machines in order to reduce computational load and network usage.
        
        All client connections after the first follow the same processing pattern, primarily handling the behavior control and synchronization of NPC vehicles.
        Since these machines carry a lighter processing load compared to the host and the first client machine, allocating additional object placement and processing tasks to the second and subsequent clients proves more efficient.
        In the evaluation section, multiple LiDAR sensors are mounted on the second client to assess the maximum number of LiDAR units that can be supported.
        To account for the impact of communication bandwidth, an additional machine is introduced, where Rviz is launched to receive and visualize the point cloud data.

    \subsection{Motivation for Selecting AWSIM}\label{subsec:DesignChoice}
        \begin{table}[t]
            \centering
            \caption{Comparison of major simulators}
            \vspace{-2mm}
            \begin{tabularx}{\columnwidth}{|X|c|c|c|}
                \hline &  3D Data Quality & System Requirements & \makecell{Large-scale \\ Simulation} \\ \hline
                    CARLA &  High $\nearrow$ & High $\nearrow$ & Low $\searrow$ \\ \hline
                    SUMO  &  $\times$ & Low $\searrow$ & High $\nearrow$ \\ \hline
                    AWSIM &  Middle $\rightarrow$ & Middle $\rightarrow$ & Middle $\rightarrow$ \\ \hline
            \end{tabularx}
            \label{tab:Comparison_simulator}
            \vspace{-5mm}
        \end{table}
        
        The Reasons for the adoption of AWSIM as an autonomous driving simulator include the ability to improve scalability via NGO, the capability for scenario-based simulation, and moderate computational resource requirements compared with others~\cite{SurveySimulator}.
        A comparison of the features of CARLA and SUMO is presented in Table~\ref{tab:Comparison_simulator}.
        The features of CARLA include the capability to perform high-quality three-dimensional rendering and generate high-quality sensor data. 
        High computational resource requirements owing to three-dimensional rendering and focus on the detailed simulation of individual vehicles render CARLA unsuitable for large-scale traffic simulation.
        
        The features of SUMO (Simulation of Urban MObility), renowned as a traffic simulator, include the capability to simulate complex traffic networks efficiently, and at large scale, as well as considerably lower computational resource requirements compared with others. 
        Inherent characteristics of traffic simulation preclude the acquisition of sensor data, and the reproduction of the physical behavior of individual vehicles is not conducted. 
        These characteristics prove suboptimal for the purpose of generating the urban-scale Dynamic Map.
        
        In that regard, AWSIM possesses the features that align with the objective. 
        Although it does not match CARLA, AWSIM is capable of high-quality three-dimensional rendering and offers flexible scenario-based customization. 
        Regarding the scalability, the simulation of traffic networks at scale comparable to SUMO proves challenging, a limitation addressed by realization of distributed processing across multiple machines via NGO. Considering these factors, the implementation of D-AWSIM was undertaken and utilized for the generation of the Dynamic Map.

\section{Evaluation}\label{sec:Evaluation}
    \begin{table}[tb]
        \centering
        \caption{Configurations of machines for Evaluation}
        \vspace{-2mm}
        \begin{tabular}{|c|c|c|c|}
            \hline
            ~   & Machine 1 & Machine 2 & Machine 3 \\ \hline
             OS & Ubuntu 22.04 & Ubuntu 22.04 & Ubuntu 22.04 \\ \hline
             CPU & \makecell[c]{Intel(R) Core(TM) \\ i7-13700KF \\ 16 core 32 thread} & 
                    \makecell[c]{Intel(R) Core(TM) \\ i9-14900K \\ 24 core 32 thread} & 
                    \makecell[c]{AMD Ryzen \\ Threadripper 7960X \\ 24 core 48 thread}\\ \hline
             RAM & 144 GB & 202 GB & 217 GB \\ \hline
             GPU & \makecell[c]{Nvidia GeForce \\ RTX 4090} &
                    \makecell[c]{Nvidia GeForce \\ RTX 4090} & 
                    \makecell[c]{Nvidia RTX 6000 \\ Ada Generation} \\ \hline
        \end{tabular}
        \label{tab:machines}
        \vspace{-5mm}
    \end{table}
    
    To demonstrate the effectiveness of D-AWSIM, evaluations are conducted from multiple perspectives.
    First, to evaluate the scalability of D-AWSIM, the number of machines is varied from one to three, and the maximum number of vehicle data streams that can be transmitted to and processed by the Dynamic Map server is measured.
    The vehicle data includes information such as the position, speed, and orientation of each vehicle.
    Each machine aggregates all vehicle data into a single stream, transmitting the data to the Dynamic Map server.
    Considering use cases, each machine should extract different subsets of data and be assigned corresponding processing tasks to make the efficient use of computational resources and support larger-scale simulations.
    However, since this evaluation focuses solely on the proposed framework, performance measurements are carried out under intentionally inefficient, high-load conditions.
    
    % Second, to show that infrastructure sensors can be flexibly installed and utilized in diverse traffic environments, the system is tested by measuring how many LiDAR sensors can be installed on the third machine when three machines are connected.
    
    Second, as a use case demonstrating interoperability with external applications, the generated Dynamic Map is integrated with Autoware~\cite{autoware}.
    The specifications of the machines used for the evaluation are shown in Table~\ref{tab:machines}, and all machines are connected via wired LAN.
    In the following sections, the machine that serves both as a server and a client is referred to as the host machine, the first connected machine as Client~1, and the next connected machine as Client~2.

    \subsection{Execution Time and Interval Time between D-AWSIM and Autoware}\label{subsec:exec_interval_time}
        \begin{figure*}[t]
            \centering
            \begin{subfigure}[t]{0.32\linewidth} % 図の幅を調整
                \centering
                \includegraphics[width=\linewidth]{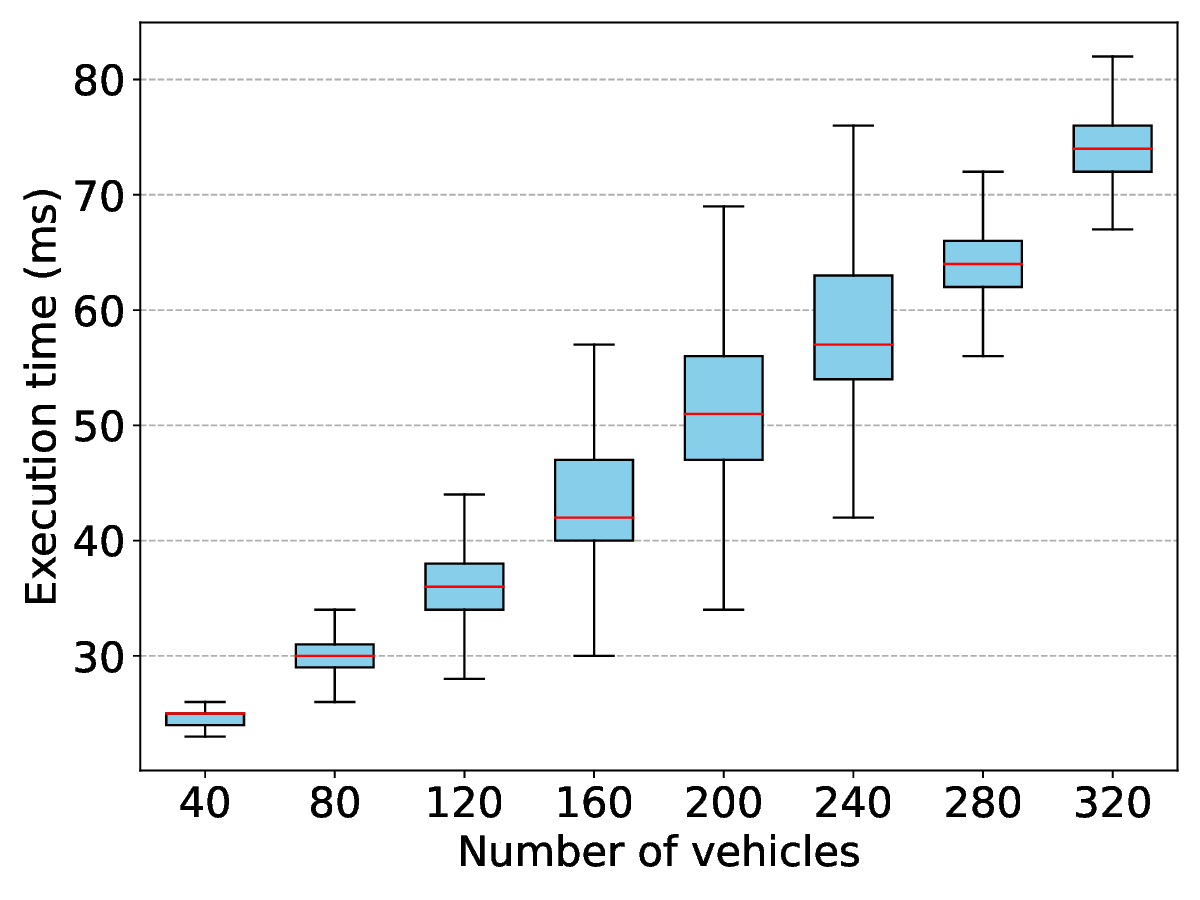}
                % \vspace{2mm}
                \caption{Host Machine.}\label{exec_1}
            \end{subfigure}
            \begin{subfigure}[t]{0.32\linewidth}
                \centering
                \includegraphics[width=\linewidth]{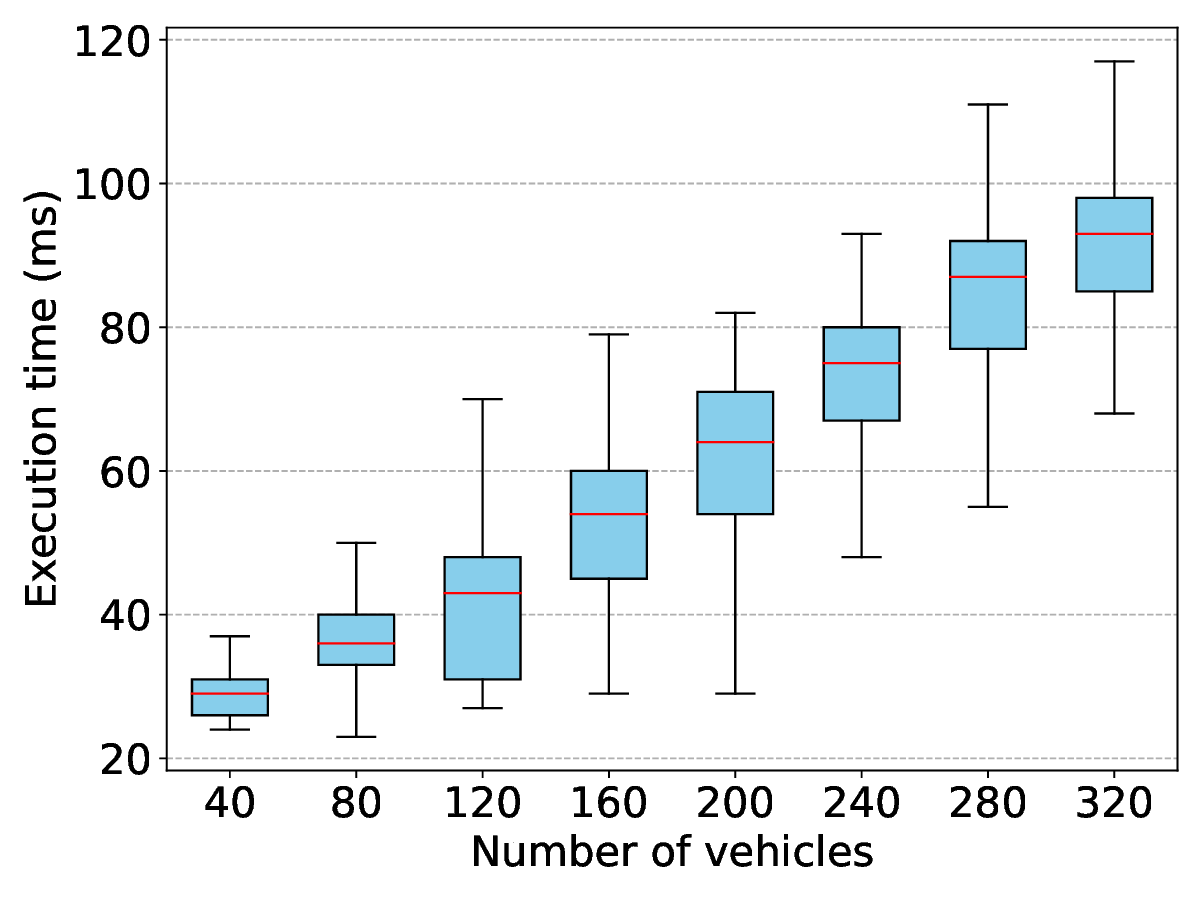}
                % \vspace{2mm}
                \caption{Client 1.}\label{exec_2}
                \vspace{-2mm}
            \end{subfigure}
            \begin{subfigure}[t]{0.32\linewidth}
                \centering
                \includegraphics[width=\linewidth]{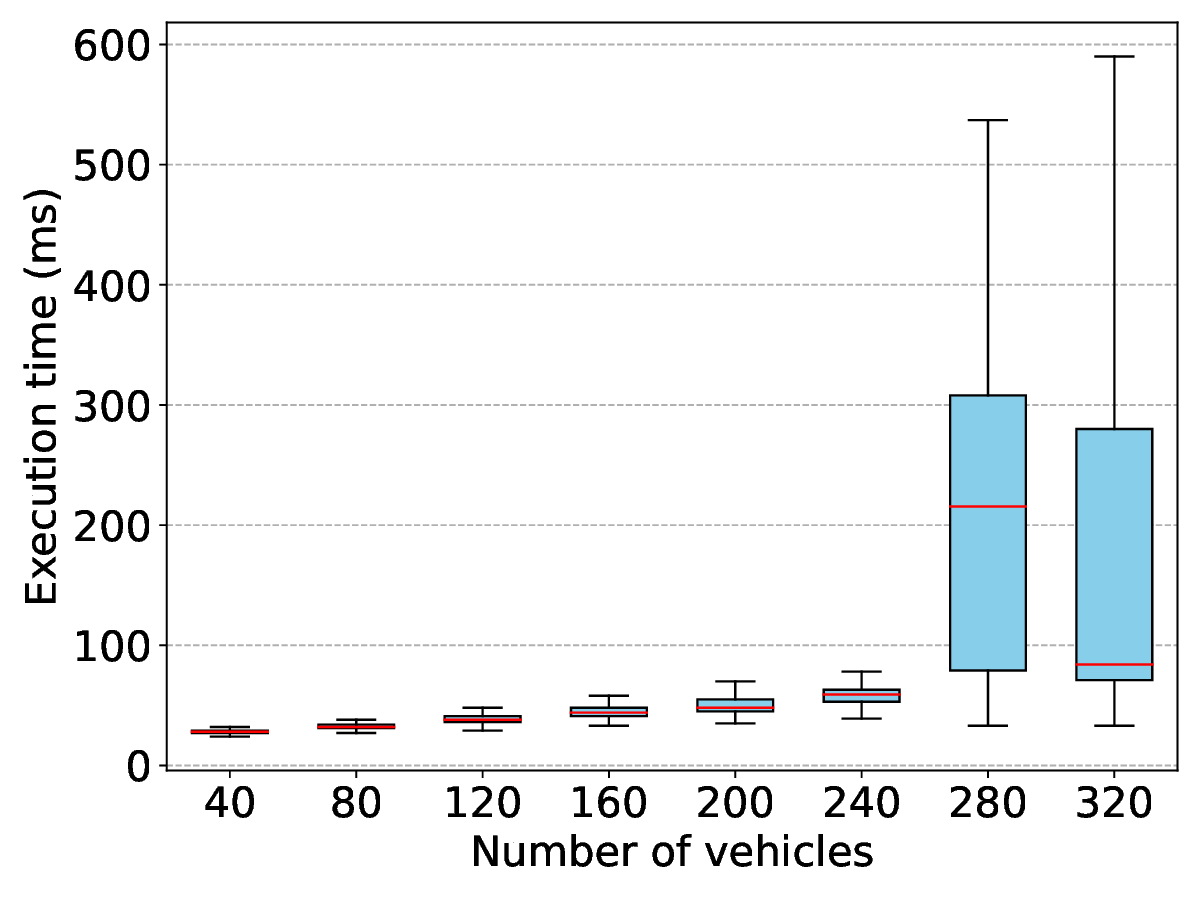}
                % \vspace{2mm}
                \caption{Client 2.}\label{exec_3}
            \end{subfigure}
            % \vspace{2mm}
            \caption{End-to-end execution time.}\label{execution}
            \vspace{-4mm}
        \end{figure*}
        \begin{figure*}[t]
            \centering
            \begin{subfigure}[t]{0.32\linewidth} % 図の幅を調整
                \centering
                \includegraphics[width=\linewidth]{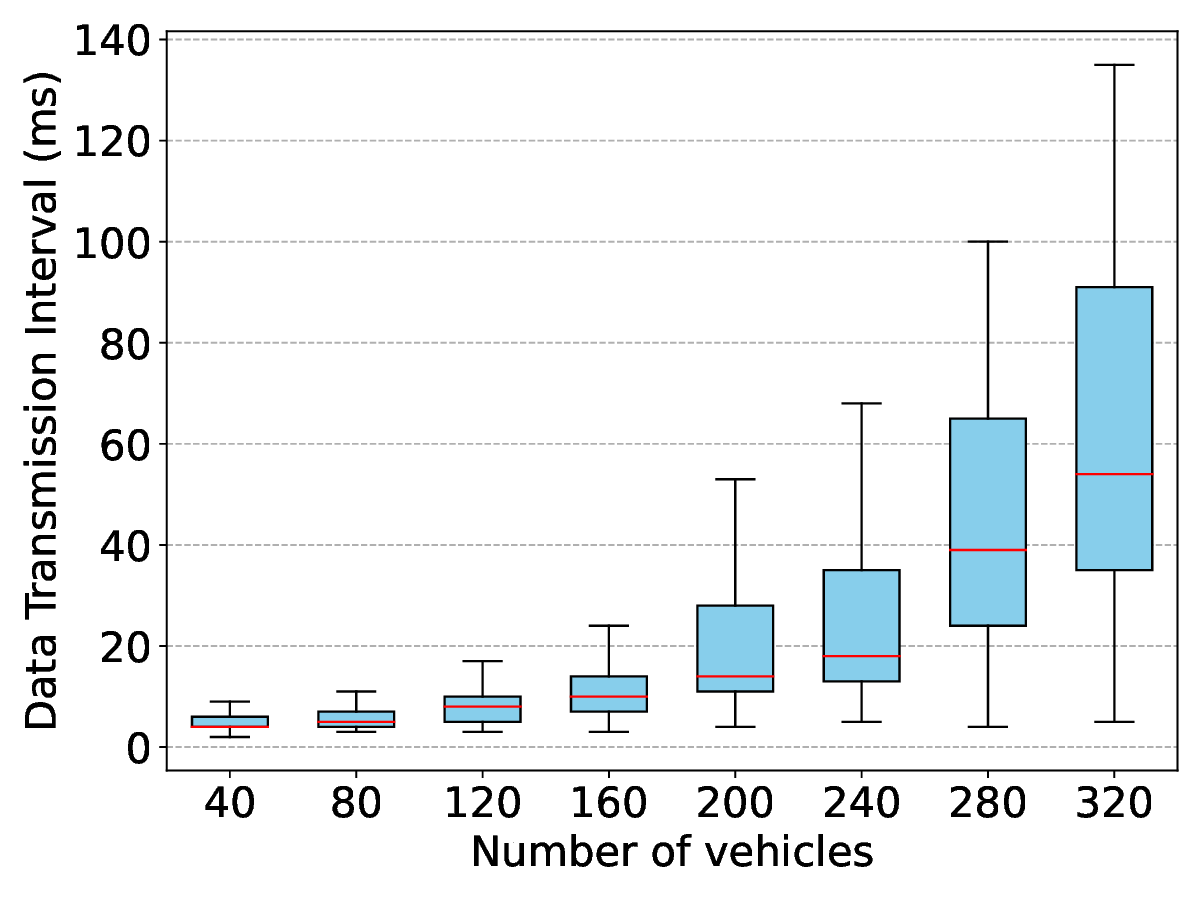}
                % \vspace{3mm}
                \caption{Host Machine}\label{interval_1}
            \end{subfigure}
            \begin{subfigure}[t]{0.32\linewidth}
                \centering
                \includegraphics[width=\linewidth]{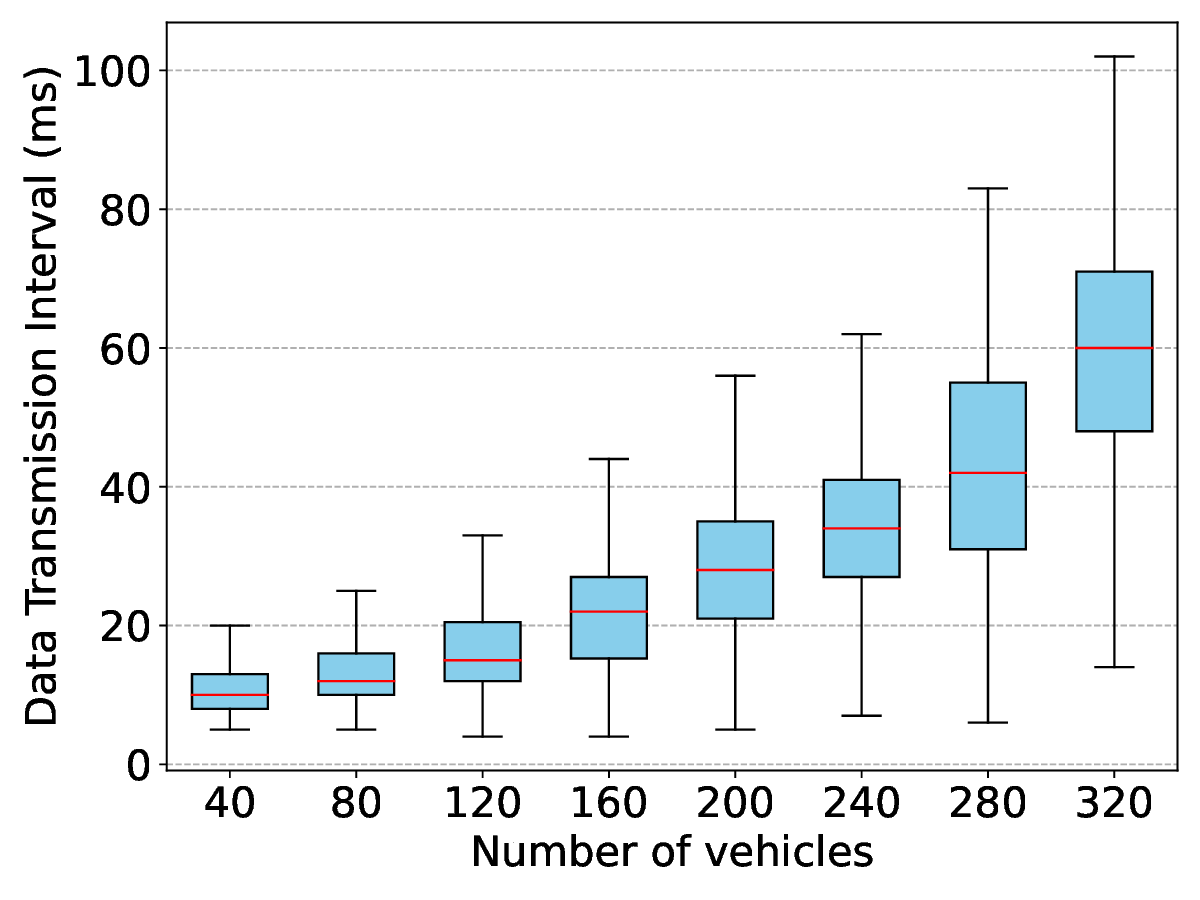}
                % \vspace{2mm}
                \caption{Client 1}\label{interval_2}
            \end{subfigure}
            \begin{subfigure}[t]{0.32\linewidth}
                \centering
                \includegraphics[width=\linewidth]{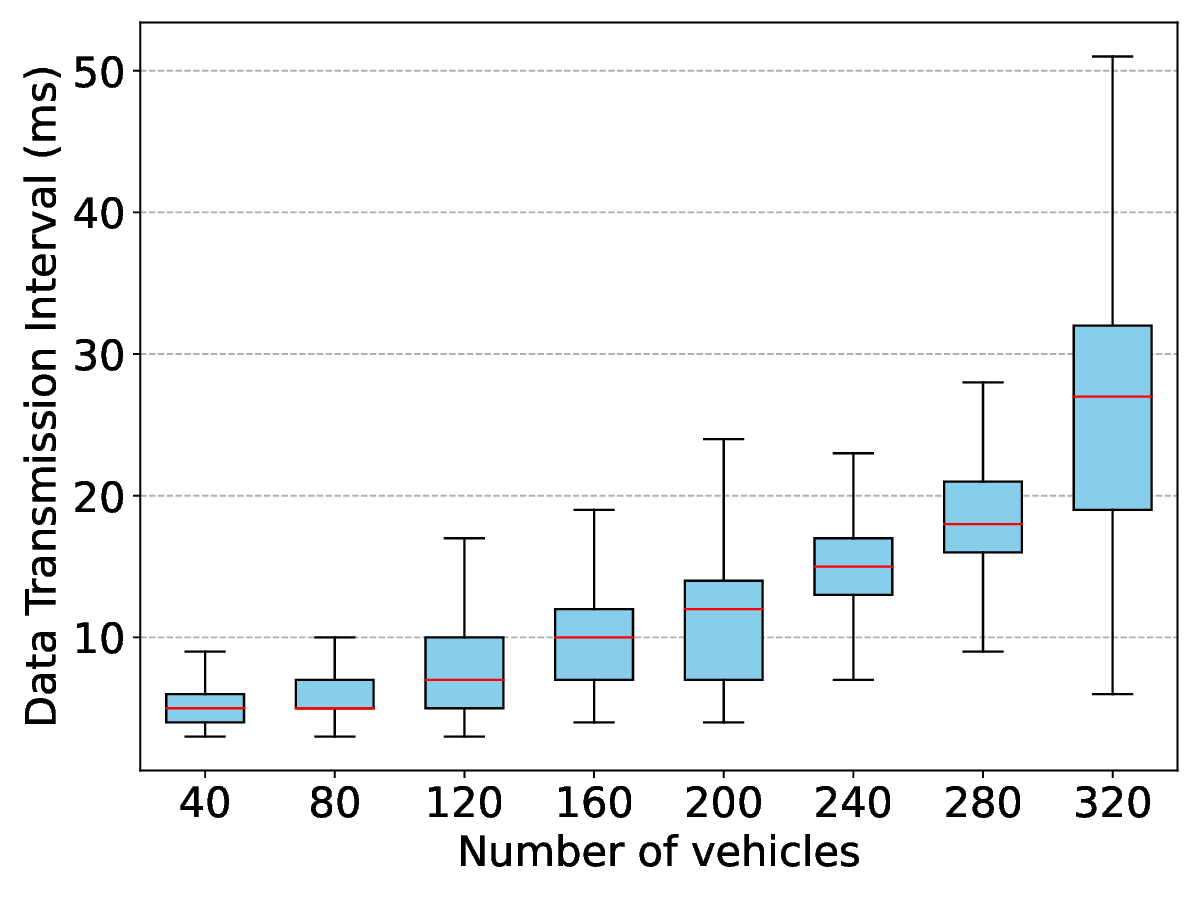}
                % \vspace{2mm}
                \caption{Client 2}\label{interval_3}
            \end{subfigure}
            \caption{Data transmission interval.}\label{interval}
            \vspace{-6mm}
        \end{figure*}
        For end-to-end execution time evaluation, the starting point is the moment when simulation data is extracted within D-AWSIM, and the end point is when the external application Autoware (based on ROS~2, Robot Operating System) receives the data.
        During this process, stream processing is performed on the Dynamic Map server, along with the conversion from JSON to ROS~2 messages.
        
        The allocation of vehicles to each machine is determined based on the processing tasks assigned to them.
        The host machine serves as both the server and the processing unit for the LiDAR sensor attached to the ego vehicle in the simulation, and Client~1 handles the processing of the camera mounted on the ego vehicle.
        Both sensors operate at 10 Hz and publish data as ROS~2 topics.
        Camera processing typically demands more computation than LiDAR processing, and its data output is about three to four times larger.
        Although these topics are not used in the time measurement evaluations, the sensor processing is still acquired within D-AWSIM.
        Therefore, these processing tasks are distributed between the host and the client machines.
        Client~2 does not handle any ego-related tasks and is responsible only for NPC vehicle control.
        Considering these processing loads, the processing related to NPC vehicles is allocated to the host machine, Client~1, and Client~2 in a 1:1:2 ratio.
        The evaluation begins with a vehicle distribution of 10:10:20 and increases the total number of vehicles by 40 for each subsequent test.
        
        The target end-to-end execution time is set to 100 ms, following the ETSI definition~\cite{LDM_etsi}.
        This value represents the maximum permissible communication latency for V2X over 5G networks and is deemed appropriate for the present metric.
        For vehicle count target, the prior study~\cite{SIM-LDM} demonstrated the ability to process up to 120 vehicles on a single machine while satisfying this timing requirement. 
        Although this study employs three machines, inter-machine communication overhead and bandwidth limitations are taken into account, leading to a doubled target of 240 vehicles. 
        This value represents a large-scale scenario for a 3D-capable autonomous driving simulator.
        
        The results for each machine are presented in Fig.~\ref{execution}~\subref{exec_1} through Fig.~\ref{execution}~\subref{exec_3}. 
        For the host machine, the 100 ms threshold remains satisfied even when the vehicle count is increased up to 320. 
        Client~1 begins to exceed 100 ms at 280 vehicles, and approximately 50\% of cases surpass the threshold at 320 vehicles. 
        Client~2 shows a marked increase at 280 vehicles, with over half of the cases exceeding the threshold.
        The results indicate that, in terms of execution time, up to 240 vehicles could be extracted.

        Next, the data transmission intervals within D-AWSIM are evaluated. 
        D-AWSIM treats data collection, data formatting, and transmission to the Dynamic Map server as a single cycle, proceeding immediately to the next cycle once the previous transmission completes. 
        In this evaluation, the interval between cycles is defined as the transmission interval, which is the time from the start of one data collection to the start of the next.
        
        The target value for this evaluation, as with execution time, is the 10 Hz (100 ms) transmission interval defined by ETSI for CAM~\cite{etsi_cam_2019} and DENM~\cite{etsi_denm_2019}. 
        CAM and DENM are standard messages used to share vehicle-to-vehicle (V2V) and vehicle-to-infrastructure (V2I) information, such as vehicle and environmental data, and support cooperative driving. 
        This metric is employed to assess the transmission interval.
        
        Fig.~\ref{interval}~\subref{interval_1} through Fig.~\ref{interval}~\subref{interval_3} present the evaluation results for each machine.
        For the host machine, the performance meets the 100 ms threshold up to 280 vehicles.
        At 320 vehicles, approximately 20\% of cases exceed the threshold.
        Client~1 likewise satisfies the threshold through 280 vehicles, with only a few cases slightly exceeding 100 ms at 320 vehicles.
        Client~2 achieves a maximum interval of 50 ms at 320 vehicles.

        The overall results indicated that up to 280 vehicles satisfied the ETSI standards. 
        The processing load can be roughly divided into three parts: data extraction within Unity, computation on DSMS, and ROS~2 node operations such as publishing.
        The measurement of the processing time for each part showed that the time from data extraction within Unity until the data reached DSMS is between 1 ms and 2 ms. 
        This does not increase significantly with the increase in the number of vehicles and is not a significant bottleneck. The processing on DSMS requires between 3 and 6 ms.
        While a slight proportional relationship is observed with the increase in the number of vehicles, the processing on DSMS does not constitute a significant bottleneck. 
        Finally, the part related to ROS~2 is identified as the major bottleneck, as the majority of the processing time is associated with ROS~2 communication and related processes. 
        This value accounted for over 90\%, calculated by subtracting the two aforementioned processing times from the E2E processing time. 
        This bottleneck primarily arises because both the processing within Unity and on DSMS are performed on the same machine, resulting in negligible data transmission and reception time. 
        Conversely, as Autoware, a set of ROS~2 nodes, runs on a separate machine, considerable time is spent on data transmission and reception. 
        This difference in deployment is the main reason for the observed bottleneck.

    \subsection{Number of Usable LiDARs}
        \begin{figure}
            \centering
            \vspace{-1mm}
            \includegraphics[width=0.95\linewidth]{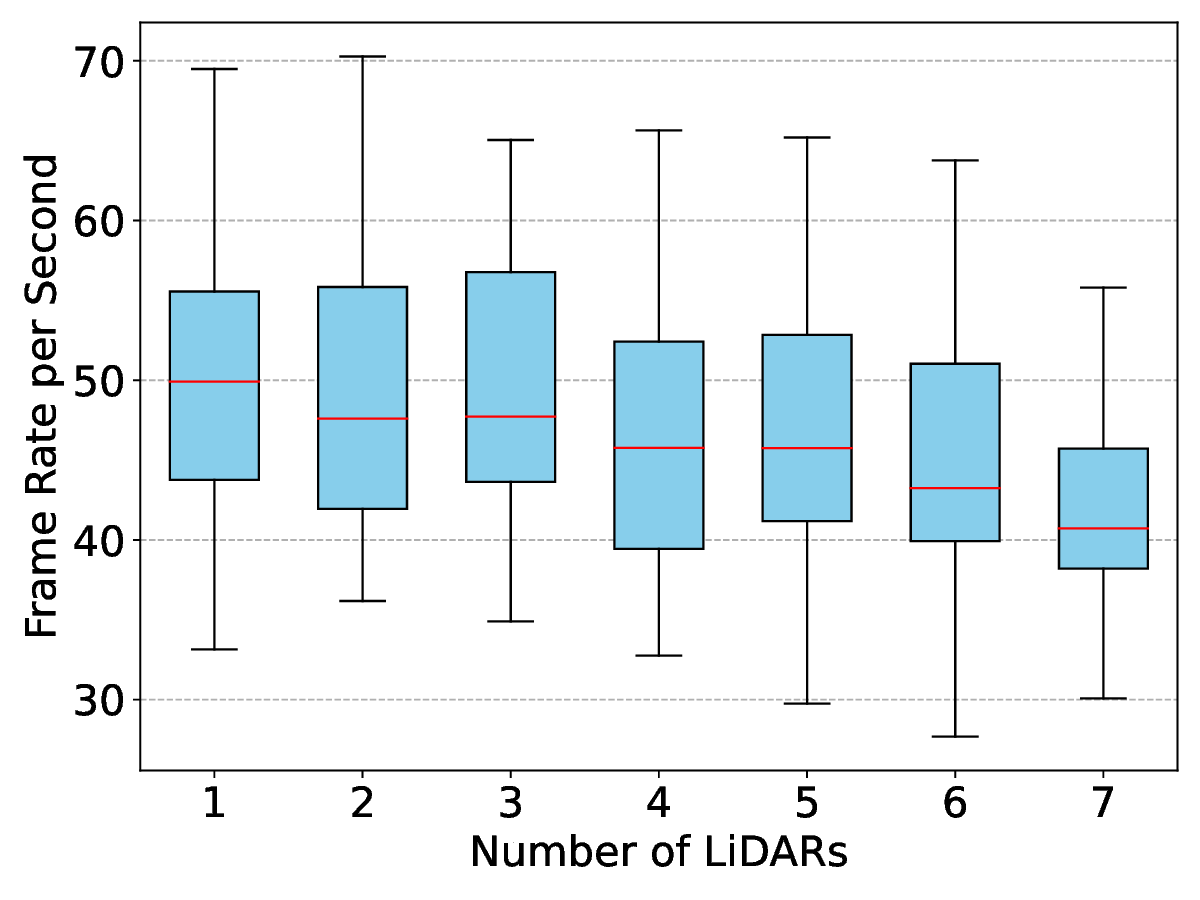}
            \vspace{-3mm}
            \caption{The number of LiDARs and FPS on Client 2}
            \label{fig:lidar}
            \vspace{-4mm}
        \end{figure}
    
        An evaluation is performed to determine how many LiDARs D-AWSIM can support. Based on the results from Section~\ref{subsec:exec_interval_time}, the total number of generated vehicles is set to 200 (allocated as 50, 50, and 100) to allow a sufficient margin relative to the performance criteria.
        All LiDARs are mounted on Client~2, and each LiDAR is installed at an intersection.
        Starting with two units, one additional LiDAR is added at each step, and the frame rate of Client~2 is recorded.
        Although a simulator frame rate of 60 Hz or higher is ideal, practical considerations, such as sensor accuracy and computational load, suggest a realistic target range of 30 to 40 Hz. 
        Accordingly, the benchmark for this evaluation is defined as maintaining a frame rate of at least 30 Hz. 
        The LiDAR model used in this evaluation is Velodyne VLP-16, publishing point clouds at 10 Hz. 
        Point cloud data received by Client~2 are forwarded to a fourth machine for visualization in Rviz.
        
        The frame rate of Client~2 trend is shown in Fig.~\ref{fig:lidar}. 
        Once eight LiDARs are installed, the average frame rate dropped to around 40 Hz, with minimum values falling to about 30 Hz. 
        These results demonstrate that Client~2 can support up to eight LiDARs while satisfying the evaluation criterion.

        The variability in frame rate is due to the synchronization timing with the host, the timing of LiDAR point cloud publication, and the transmission of vehicle data to the Dynamic Map server, which can generate temporary spikes in processing load that coincide with frame rate measurement. 
        Additionally, because LiDAR processing load fluctuates with the number of laser returns, variations in traffic density at the installation site also affect processing time. 
        These fluctuations in processing load further contribute to the observed frame rate variability.
                
    \subsection{Demonstration of Connecting Dynamic Map to Autoware}
        \begin{figure}[t]
            \centering
            \begin{subfigure}[t]{0.49\linewidth}
                \centering
                \includegraphics[width=\linewidth]{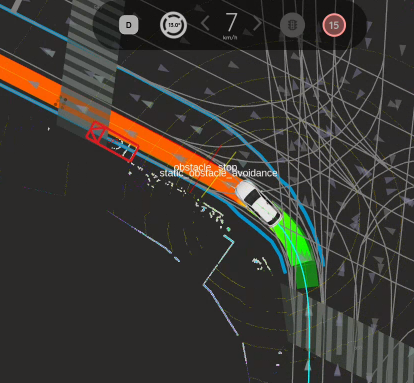}
                % \vspace{3mm}
                \caption{Without Dynamic Map.}\label{withoutDM}
            \end{subfigure}
            \begin{subfigure}[t]{0.49\linewidth} % 図の幅を調整
                \centering
                \includegraphics[width=\linewidth]{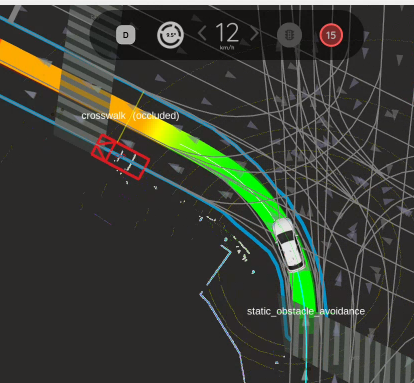}
                % \vspace{3mm}
                \caption{With Dynamic Map.}\label{withDM}
            \end{subfigure}
            \vspace{-1mm}
            \caption{Autoware connected to Dynamic Map.}\label{demo}
            \vspace{-5mm}
        \end{figure}
        This section presents a use case in which a Dynamic Map generated from simulation data is connected to Autoware~\cite{autoware}.
        This may be an example of a use case for a Dynamic Map while also showing how the generated Dynamic Map may be connected to various external applications.        
        Since Autoware is a ROS~2-based automated driving software, converting the data returned from the dynamic map into ROS~2 messages becomes necessary.
        Therefore, a simple converter is created to extract the necessary data from the data returned in JSON format and convert it into the necessary ROS~2 messages, and the output was sent to Autoware.
        
        The experiment assumes a scenario in which an ego vehicle makes a left turn at an intersection with poor visibility, and one vehicle is stopped as an obstacle at the end of the left turn.
        Under this scenario, the effect of the linkage between Autoware and the dynamic map on the route planning of the ego vehicle is investigated.
        
        The execution of the assumed scenario is illustrated in Fig.~\ref{demo}\subref{withoutDM} and Fig.~\ref{demo}\subref{withDM}.
        When the ego vehicle enters the intersection without integration with the Dynamic Map, no avoidance route is generated upon recognition of the stopped vehicle as shown in Fig.~\ref{demo}\subref{withoutDM}.
        As the vehicle continues forward, avoidance of the obstacle fails, resulting in a halt behind the stopped vehicle.
        In contrast, when entering the intersection with the Dynamic Map integration, a route that accounts for the stopped vehicle is generated in advance as shown Fig.~\ref{demo}\subref{withDM}.
        As a result, the ego vehicle is able to smoothly avoid the obstacle and continue its path without interruption.
        
        These results show that, even in simple scenarios, the lack of information about blind spots poses a significant risk and has a major impact on safety.
        They also demonstrate that the Dynamic Map serves as a promising approach in such situations, offering a considerable improvement in safety.

\section{Lessons Learned}\label{sec:lesson_learned}
    The insights and challenges gained through the implementation and evaluation of this study are organized below, and concrete recommendations for future research are presented.
    
    The first consideration is to minimize bandwidth congestion and avoid creating bottlenecks. 
    Although each data payload is small, simulating a large-scale traffic environment requires hundreds of vehicles. 
    Moreover, because data transmissions occur multiple times per second, even slight increases in data volume directly affect overall performance. 
    In D-AWSIM, only the vehicle data generated by each client, limited to essential parameters such as position and speed, is sent to the server, thereby reducing data volume and cutting communication latency.
    
    The second consideration involves distributing processing tasks appropriately. 
    In AWSIM, each NPC vehicle is controlled by a script attached to its own client. 
    Simply synchronizing all vehicle data would cause vehicles controlled by one client to be re-controlled locally by every other client.
    To prevent this, D-AWSIM distinguishes between vehicles generated by its own client and those generated by other clients, and applies control logic only to its own-generated vehicles.
    Without this differentiation, the number of vehicles D-AWSIM could handle dropped to less than half.
    
    The third consideration is designing for anticipated communication delays during data synchronization. 
    As noted earlier, delays can occur due to high data volumes or processing loads. 
    Simply applying synchronized data to update the simulation environment can cause object jitter when packets arrive late. 
    To mitigate this, the system must preserve objects for a fixed number of frames if no new data arrives. 
    In D-AWSIM, a threshold of 10 frames is used.

    A key insight from our evaluation is the significance of distributed simulation in overcoming scalability limitations of single-host systems. 
    Prior work such as SIM-LDM~\cite{SIM-LDM}, which relies on a single simulator instance, supports up to 120 vehicles due to rendering and computation limits.
    In contrast, the distributed architecture of D-AWSIM achieved smooth operation with up to 240 vehicles, while maintaining end-to-end latency and transmission interval within the target thresholds. 
    This highlights that offloading computation and carefully managing inter-machine communication are essential for realizing realistic, large-scale simulation environments.
    
    The primary limitation lies in the lack of optimization across all aspects of the system. 
    For example, D-AWSIM currently employs a static distribution of processing tasks; adapting this distribution dynamically based on each resource utilization of the client and network conditions would enable more efficient use of available resources.

\section{Related work}\label{sec:Related}
    \begin{table}[t]
        \centering
        \caption{Comparison of the proposed framework with existing studies}
        \vspace{-1mm}
        \label{tab:related_work}
        { % ← ここから局所グループ開始
            \renewcommand{\arraystretch}{1.0}
            \begin{tabularx}{\columnwidth}{|X|c|c|c|c|} \hline
                ~                               & \textbf{Open-source} & \textbf{UEA}  & \textbf{CGD} &  \textbf{Scalability} \\ \hline
                AutoC2X~\cite{autoC2X}          & \checkmark & ~            &  ~            & ~             \\ \hline
                AutowareV2X~\cite{autoware_v2x} & \checkmark & \checkmark   &  ~            & ~    \\ \hline
                SIM-LDM~\cite{SIM-LDM}          & \checkmark & \checkmark   & \checkmark    & ~    \\ \hline
                Sky-Drive~\cite{sky-drive}      & Upcoming   & \checkmark   &  ~            & \checkmark    \\ \hline
                % \makecell[l]{HLA \\Co-Simulations~\cite{HLA}}& ~            & \checkmark    & ~  & \checkmark  \\ \hline
                HLA~\cite{HLA}& ~            & \checkmark    & ~  & \checkmark  \\ \hline
                EDRP and LDMP~\cite{offload}    & ~          & \checkmark   &  ~            & \checkmark    \\ \hline
                iLDM~\cite{iLDM}                & ~          & ~            & \checkmark    & \checkmark     \\ \hline
                LiveMap~\cite{liveMap}          & ~          & ~            & \checkmark    & \checkmark     \\ \hline
                Proposed Framework              & \checkmark & \checkmark   & \checkmark    & \checkmark    \\ \hline
            \end{tabularx}
        }
        \begin{minipage}{\textwidth}
            \vspace{1mm}
            \raggedright % 左揃え
            UEA: Usable with External Applications \\
            CGD: Capable of Generating Dynamic Map 
            \vspace{-3mm}
        \end{minipage}
    \end{table}
    This section introduces the work related to this paper.
    The comparison results with existing studies are shown in Table~\ref{tab:related_work}.
    AutoC2X~\cite{autoC2X} is a system based on OpenC2X~\cite{openC2X}, enabling cooperative perception for standalone autonomous vehicles based on Autoware.
    AutoC2X allows autonomous vehicles to share detected objects with surrounding vehicles and infrastructure sensors.
    AutowareV2X~\cite{autoware_v2x}, which provides external connectivity to Autoware, has similar functionalities, and the demonstration has shown that AutowareV2X can perform evasive actions based on shared information.
    
    SIM-LDM~\cite{SIM-LDM} is a study on the automatic generation of a Local Dynamic Map using a simulator, and also supports processing queries dynamically.
    However, One major difference is that SIM-LDM uses a single simulator, which lacks scalability.
    Therefore, as the number of vehicles increases, the frame rate of the simulator drops significantly.
    As a result, the number of vehicles that could be extracted after meeting the requirements was limited to 120 by SIM-LDM.
    Sky-Drive~\cite{sky-drive} is a distributed multi-agent simulation based on CARLA, which is an Unreal Engine simulator.
    This system, similar to D-AWSIM, allows a theoretical connection to any number of machines and achieves this through RPC-based extensions.
    Although an open-source release is planned, the system is not publicly available at the time of writing, and no information has been provided regarding the maximum number of vehicles supported.
    However, considering the specifications required by CARLA and its ability to generate higher quality data than AWSIM, D-AWSIM is more suitable for applications.
    HLA Co-Simulations~\cite{HLA} combines a single virtual federate that manages multiple vehicles internally and compresses connections to runtime in order to integrate a traffic flow simulator and a communication network simulator.

    While relational database-based LDM are mainstream, iLDM~\cite{iLDM} proposes a graph-based LDM to address the limitations in scalability and flexibility of RDBMS.
    LiveMap~\cite{liveMap} is a framework designed for constructing a Dynamic Map in edge computing environments.
    However, the edge server lacks the extended capabilities for processing queries  dynamically from external applications, as seen in D-AWSIM.
    The integrated platform for computation offloading and dynamic data sharing enables rapid sensor data processing and information sharing essential for cooperative driving~\cite{offload}.
    The Edge Dynamic Map architecture~\cite{EDM_architecture_for_C-ITS}, combining Multi-access Edge Computing and a time-series database, proposes the EDM architecture to enhance C-ITS by leveraging MEC and 5G technologies.
    
    First Mile~\cite{First_Mile} is an open innovation lab equipped with 89 advanced sensors installed along a 3.5 km public road section to contribute to the acquisition of practical insights.
    This lab provides a variety of communication interfaces, allowing researchers and companies to conduct experimental validations freely,
    and also plays a role in supporting the construction of digital twins and offering high-quality real-world datasets.

    As shown above, research on a Dynamic Map, distributed simulation, and experimental validation is actively being pursued.
    However, no existing studies focus specifically on the generation of the Dynamic Map, despite the fact that ensuring availability is essential for its future development.
    D-AWSIM is proposed as a solution to this problem.

\section{Conclusion}\label{sec:Conclusion}
    This paper proposed a distributed simulation, D-AWSIM, and a framework for automatic Dynamic Map generation using D-AWSIM, in order to make a Dynamic Map available to researchers.
    Autonomous driving simulators are required to process many high-load processes, such as sensor data and agent control at high speed, which makes a single-simulator difficult to perform large-scale simulations on a single machine due to issues such as computational resources.
    D-AWSIM achieves processing distribution by linking simulators on multiple machines, enabling large-scale simulations that are difficult to perform on a single machine.
    In addition, a Dynamic Map, which is expected to be used in a wide range of traffic environments, is difficult for anyone to use freely due to the high cost of infrastructure sensors and regulatory issues.
    D-AWSIM is utilized to generate automatically the Dynamic Map from large-scale traffic environment simulation data.
    The evaluation results of the proposed framework showed that the system enables to handle data for up to 240 vehicles overall.
    
    Future work includes the generation of messages such as CAM and DENM, which are expected to help improve safety, using the data aggregated in the Dynamic Map, as well as the generation and handling of semi-static and semi-dynamic data such as traffic accident information and traffic signs.
    In addition, more efficient synchronization and object control algorithms may be devised and adapted to reduce latency.
    These potentially reduce synchronization delays and enable more accurate simulations.

% \vspace{-3mm}

\bibliographystyle{IEEEtran}
\bibliography{references}
\end{document}